\documentclass[10pt,twocolumn,letterpaper]{article}

\usepackage{iccv}
\usepackage{times}
\usepackage{epsfig}
\usepackage{graphicx}
\usepackage{amsmath}
\usepackage{amssymb}
\usepackage{subcaption}

\usepackage[breaklinks=true,bookmarks=false]{hyperref}

\iccvfinalcopy 

\def\iccvPaperID{****} 

\ificcvfinal\fi

\begin{document}

\title{Dynamic Multi-Task Learning for Face Recognition with Facial Expression}


\author{Zuheng Ming{$^1$} ~Junshi Xia{$^2$}  ~Muhammad Muzzamil Luqman{$^1$}  ~Jean-Christophe Burie{$^1$} ~Kaixing Zhao{$^3$}\\
{$^1$}L3i, University of La Rochelle, France \\{$^2$}RIKEN Center for Advanced Intelligence Project (AIP), RIKEN,Tokyo 103-0027, Japan \\ {$^3$}IRIT, University of Toulouse\\
{\tt\small $\textit{\{zuheng.ming, mluqma01, jcburie\}}$@univ.lr-fr},~{\tt\small $\textit{jushi.xia}$@riken.jp},~{\tt\small $\textit{kaixing.zhao}$@irit.fr}
}

\maketitle
\ificcvfinal\thispagestyle{empty}\fi

\begin{abstract}
Benefiting from the joint learning of the multiple tasks in the deep multi-task networks, many applications have shown the promising performance comparing to single-task learning. However, the performance of multi-task learning framework is highly dependant on the relative weights of the tasks. How to assign the weight of each task is a critical issue in the multi-task learning. Instead of tuning the weights manually which is exhausted and time-consuming, in this paper we propose an approach which can dynamically adapt the weights of the tasks according to the difficulty for training the task. Specifically, the proposed method does not introduce the hyperparameters and the simple structure allows  the other multi-task deep learning networks can easily realize or reproduce this method. We demonstrate our approach for face recognition with facial expression and facial expression recognition from a single input image based on a deep multi-task learning Conventional Neural Networks (CNNs). Both the theoretical analysis and the experimental results demonstrate the effectiveness of the proposed dynamic multi-task learning method. This multi-task learning with dynamic weights also boosts of the performance on the different tasks comparing to the state-of-art methods with single-task learning. \footnote{\url{https://github.com/hengxyz/Dynamic_multi-task-learning.git}}

\end{abstract}

\section{Introduction}
\begin{figure*}[t]
\begin{center}
   \includegraphics[width=0.9\linewidth]{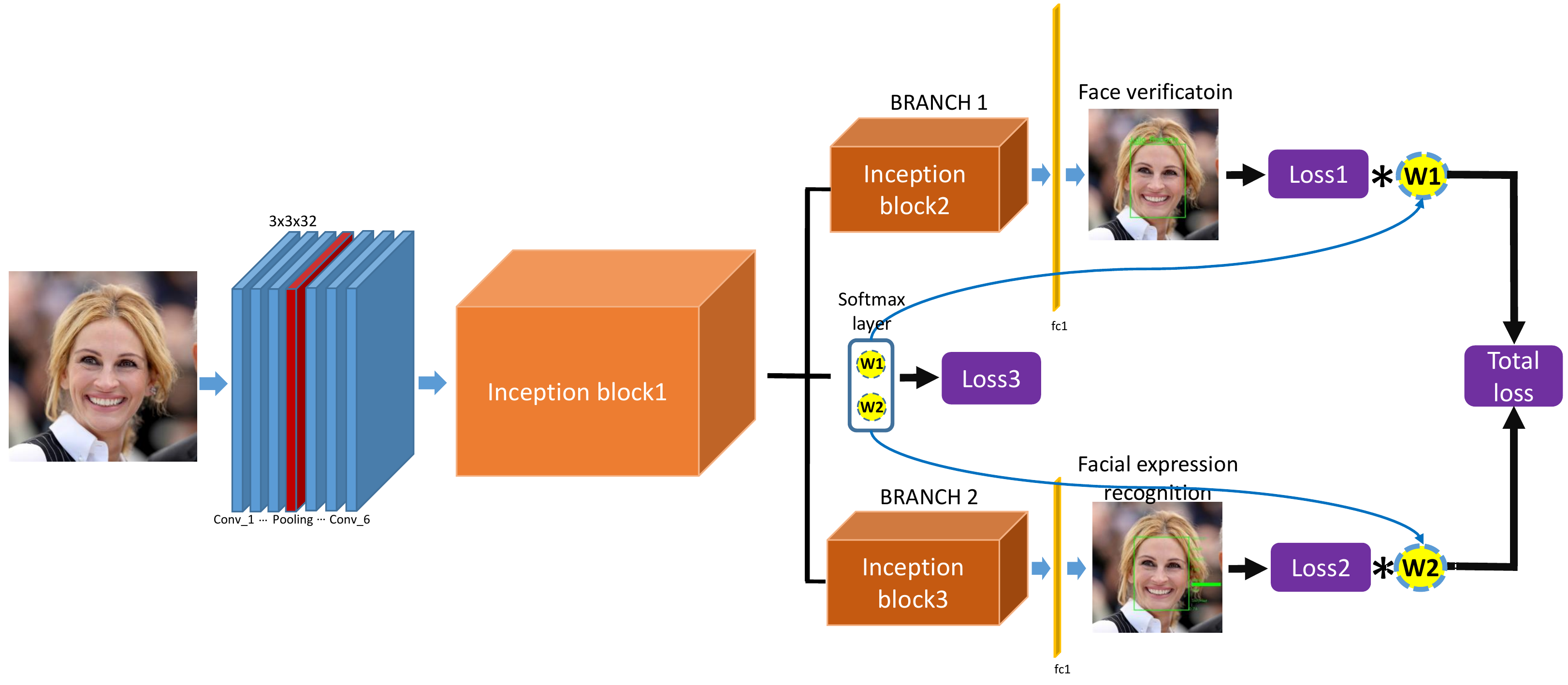}
\end{center}
   \caption{The proposed multi-task framework with dynamic weights of tasks to simultaneously perform face recognition and facial expression recognition. The dynamic weights of tasks can adapt automatically according to the importance of tasks.}
\label{fig:framework}
\end{figure*}

Multi-task learning has been used successfully across many areas of machine learning~\cite{ruder2017overview}, from natural language processing and speech recognition~\cite{collobert2008unified, deng2013new} to computer vision~\cite{girshick2015fast}. By joint learning in multiple tasks in the related domains  with different information, especially from information-rich tasks to information-poor ones, the multi-task learning can capture a representation of features being difficult learned by one task but can easily learned by another task~\cite{murugesan2016adaptive}. Thus the multi-task learning can be conducted not only for improving the performance of the systems which aims to predict multiple objectives but also can utilise for improving a specific task by leveraging the related domain-specific information contained in the auxiliary tasks. In this work, we explore the multi-learning for face recognition with facial expression. Thanks to the progress of the representing learning with the deep CNNs, face recognition has made remarkable progress in the recent decade ~\cite{taigman2014deepface, parkhi2015deep, schroff2015facenet, liu2017sphereface}. These works have achieved or beyond the human-level performance on the benchmarks LFW\cite{huang2007labeled}, YTF\cite{wolf2011face}. The challenges of face recognition such as the variation of the pose, the illumination and the occlusion have been well investigated in many researches, nevertheless face recognition for the face with the non-rigid deformation such as the ones introduced by the facial expression has not been sufficiently studied especially in the 2D face recognition domain. Some 3D based methods have been proposed to deal with this issue such as~\cite{zhu2015high, kakadiaris2007three,chang2006multiple}, in which~\cite{zhu2015high} presents the method by using the 3D facial model to normalise the facial expression and then maps the normalised face to the 2D image to employ face recognition. In order to leverage the promising progress in the numerous 2D face recognition and facial expression recognition researches particularly based on the deep neural networks, we propose to combine the face recognition task and the facial expression recognition task in the unified multi-task framework aiming to jointly learn the sharing features and task-specific features to boost the performance of each task. \figurename~\ref{fig:framework} shows the multi-task framework proposed in this work.      

\begin{figure*}
\centering
\begin{minipage}{.5\textwidth}
  \centering
  \includegraphics[width=\linewidth]{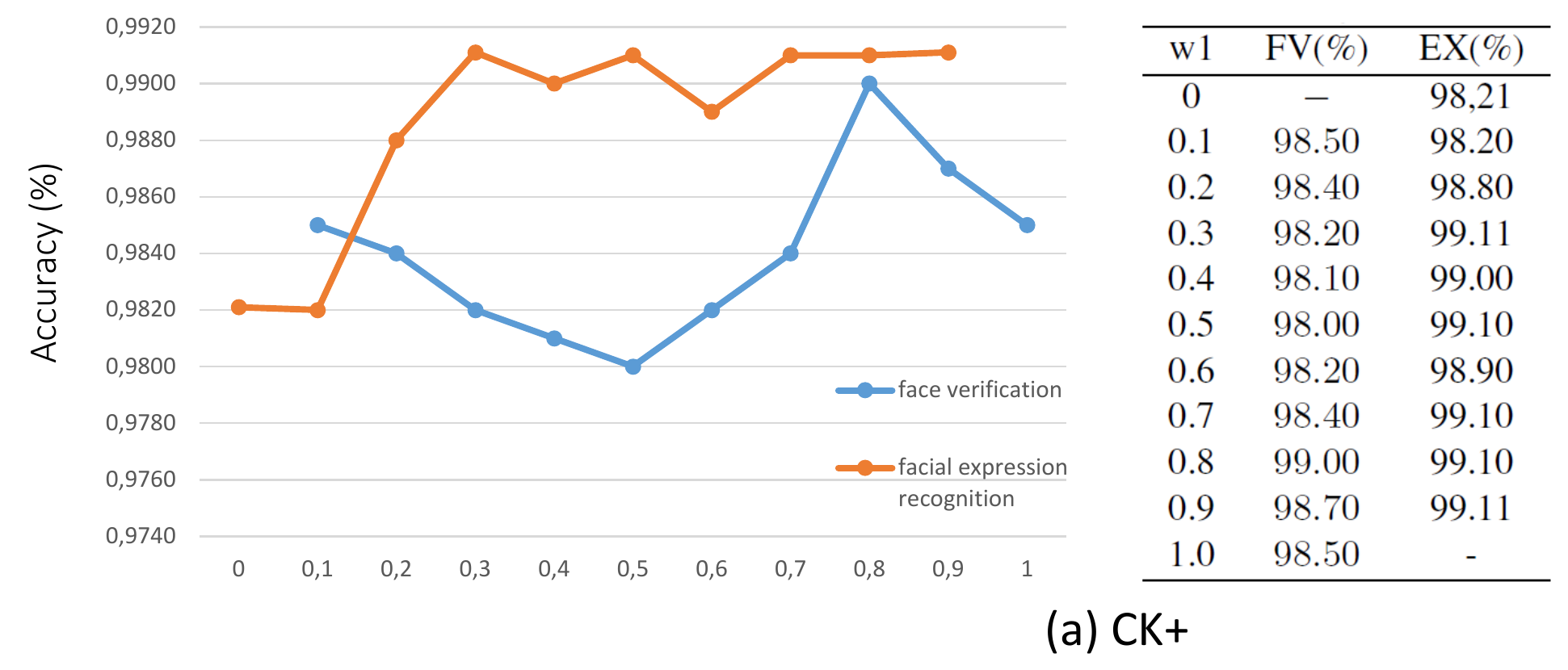}
  \label{fig:test1}
\end{minipage}%
\begin{minipage}{.5\textwidth}
  \centering
  \includegraphics[width=\linewidth]{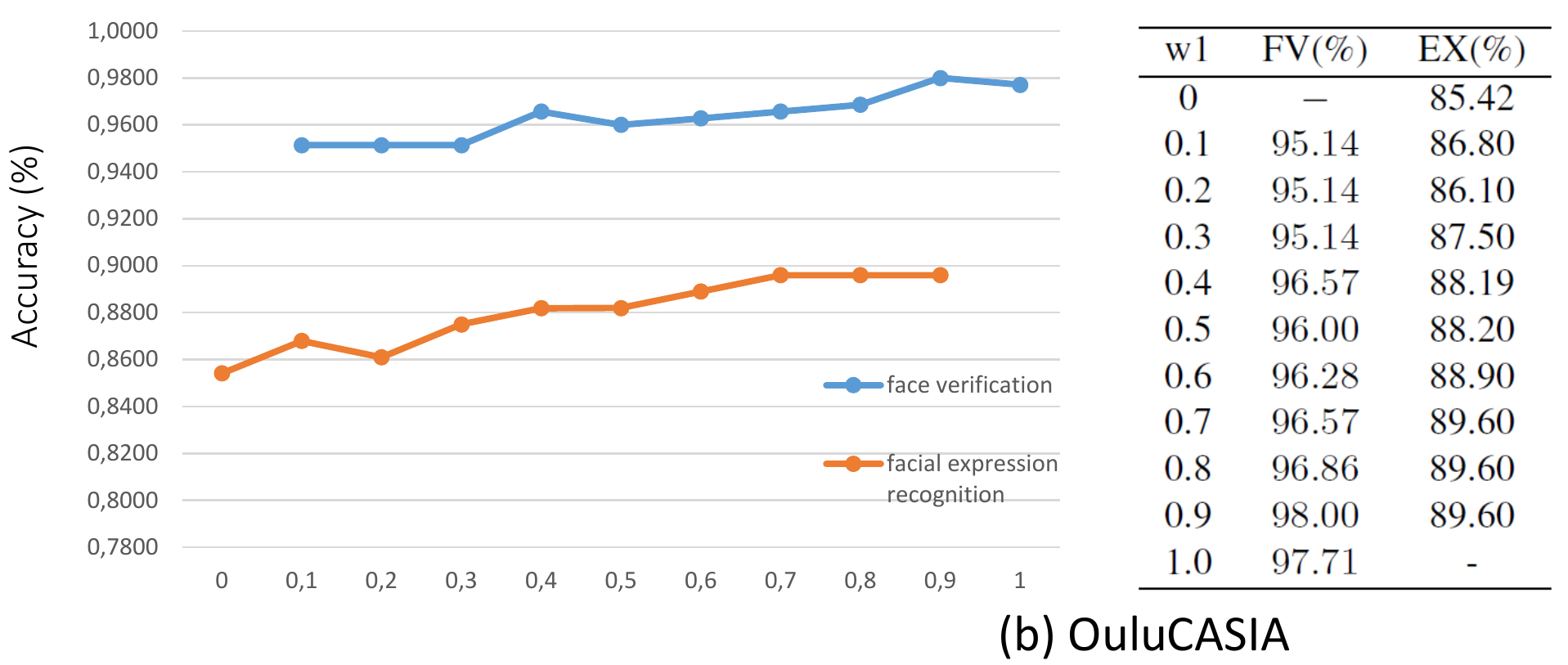}
  \label{fig:test2}
\end{minipage}
\caption{Performances of the task of face verification for facial expression images  (FV / blue) with the manually setting weight $w1$ and the task of facial expression recognition (EX / red)  on the different datasets CK+ and OuluCASIA.}
\label{fig:manualweights}
\end{figure*}

How to set the weights of tasks is a crucial issue in the multi-task learning. The weights determine the importance of the different tasks in the holistic networks. Many works simply set the equal values for all tasks or experimentally set the weights of the tasks. In ~\cite{chen2017multi}, the authors assign equal weights to the ranking task and the binary classification task for the person re-identification. However the multi-task learning is an optimization problem for multiple objectives. The main task and the side tasks with different objective have different importance in the overall loss meanwhile the difficulty of the training of each task is also different. Thus it is arbitrary to assign equal weights for tasks for multi-task learning. We also verified this point in our work by manually setting the weights of tasks from 0 to 1 with the interval of 0.1. As shown in \figurename~\ref{fig:manualweights}, either for the facial expression recognition task or for the face recognition task, the best performance are obtained with the different weights of tasks rather than the equal weights of each task.       
Most of the multi-task learning methods search the optimal weights of the tasks by the experimental methods, for instance Hyperface~\cite{ranjan2017hyperface} manually set the weights of the  tasks such as the face detection, landmarks localization, pose estimation and gender recognition according to their importance in the overall loss, and ~\cite{tian2015pedestrian} obtain the optimal weights by a greedy search for pedestrian detection tasks with the different attributes. Besides the cost of time and being exhausting, these methods setting the weights as the fix values to optimize the tasks ignore the variation of the the importance or the difficulty of the tasks during the training processing. Rather than the methods with fix weights which can be so called static weights methods, the methods~\cite{chen2017gradnorm, kendall2018multi,zhang2016learning} update the weights or part of the weights of the tasks during the training of the networks. \cite{zhang2016learning} set the weight of the main task as 1 while the auxiliary tasks are weighted by the dynamic weights $\lambda_t$ which are updated by an analytical solution. \cite{kendall2018multi} introduces a uncertainty coefficient $\theta$ to revise the softmax loss function of each task. 
Unlike these methods which need to introduce the additional hyperparameters to update the weights of tasks, we propose to use a softmax layer adding to the end of the hidden sharing layers of the multi-task networks to generate the dynamic weights of the tasks (see \figurename~\ref{fig:framework}). Each unit of this softmax layer is corresponding to a weight of a task and no more hyperparameter is introduced for updating the tasks weights. Rather than ~\cite{yin2018multi} updating simultaneously  the dynamic weights of tasks and the filters weights of the networks by the unify total loss of the networks, we propose a new loss function to update the dynamic weights which enable the networks to focus on the training of the hard task by automatically assigning a larger weight. On the contrary, ~\cite{yin2018multi} always updates the smaller weight for the hard task and the larger weight for the easy task which results the hard task is far from being fully trained and the networks stuck in the worthless training of the over trained easy task. 
This is due to use the total loss of the networks to simultaneously update the weights of tasks, in which the dynamic weights of the tasks are also in the function of the weights of networks, i.e. $\mathcal{L}_{total}(\Theta)=w_1({\Theta_0})\mathcal{L}_{1}(\Theta_1)+w_2({\Theta_0})\mathcal{L}_{2}(\Theta_2)\quad s.t.\quad w_1+w_2=1$ and $\{\Theta_0, \Theta_1,\Theta_2\}=\Theta$ are the weights of the networks. The optimization of $\Theta$ by the total loss $\mathcal{L}_{total}$ aims to decrease the total loss as much as possible, thus when the hard task has a large loss the fastest way to decrease the total loss is to shrinkage its weight $w_i$ so that the weighted loss of the hard task can be cut down rapidly. This is why the hard task always has a small task weight while the easy task has a large weight.    

In a summary, our main contributions of this paper are.
\begin{itemize}
	\item We propose a dynamic multi-task learning method which can automatically update the weight of task according to the importance of task during the training. 
    \item Both the theoretical analysis and the experimental results demonstrate the proposed dynamic multi-task learning enable to focus on the training of the hard task to achieve better efficiency and performance. 
    \item We have demonstrated that, for both face verification with facial expression and facial expression recognition tasks, the proposed multi-task learning can outperform the state-of-the-art performance on the datasets  CK+~\cite{lucey2010extended}, OuluCASIA~\cite{zhao2011facial}.  
    \item The proposed method is simple and does not introduce the hyperparameters,which can be easily realized and reproduce in the other deep multi-task learning frameworks. 

\end{itemize}

The remainder of this paper is organized as follows: Section II briefly reviews the related works; Section III describes the architecture of the dynamic multi-task network. Section IV presents the approach of multi-task learning with dynamic weights following by Section V where the experimental results are analyzed. Finally, in Section VI, we draw the conclusions and present the future works.

\section{Related works}

Multi-task learning not only helps to learn more than one task in a single network but also can improve upon your main task with an auxiliary task~\cite{ruder2017overview}. In this work, we focus on the multi-task learning in the context of the deep CNNs. According to the means of updating the weights of tasks, the multi-task learning can be divided into two categories: the static method and dynamic method. In the static methods, the weights of tasks are set manually before the training of the networks and they are fixed during the whole training of the networks~\cite{ranjan2017hyperface,girshick2015fast,  chen2017multi,yim2015rotating,tian2015pedestrian}; while the dynamic methods initialize the weights of the tasks at the beginning of the training and update the weights during the training processing~\cite{chen2017gradnorm, kendall2018multi, zhang2016learning, yin2018multi}. There are two ways for setting the weights in the static methods. The first way is to simply set the equal weights of task such as Fast R-CNN~\cite{girshick2015fast} and ~\cite{chen2017multi, yim2015rotating}. In Fast R-CNN, the author uses a multi-task loss to jointly train the classification and bounding-box regression for object detection. The classification task is set as the main task and the bounding-box regression is set as the side task weighted by $\lambda$. In the experiments the $\lambda$ is set to 1. 
The second way to set the weights is manually searching the optimal weights by the experimental methods. Hyperface~\cite{ranjan2017hyperface} proposed a multi-task learning algorithm for face detection, landmarks localization, pose estimation and gender recognition using deep CNNs. The tasks have been set the different weights according to the importance of the task. 
~\cite{chen2017multi} integrated the classification task and the ranking task in a multi-task networks for person re-identification problem. Each task has been set with a equal weight to jointly optimizing the two tasks simultaneously. 
Tian et al.~\cite{tian2015pedestrian} fix the weight for the main task to 1, and obtain the weights of all side tasks via a greedy search within 0 and 1. 
In ~\cite{chen2017gradnorm} an additional loss in function of the gradient of the weighted losses of tasks is proposed to update the weights meanwhile an hyperparameter is introduced for balancing the training of different tasks. 
~\cite{kendall2018multi} introduces a uncertainty coefficient $\theta$ to combine the multiple loss functions. The $\theta$ can be fixed manually or learned based on the total loss. 
Zhang et al.~\cite{zhang2016learning} propose a multi-task networks for face landmarks detection and the recognition of the facial attributes. The face landmarks detection is set as the main task with the task weight 1 and the tasks for recognition of the different facial attributes are set as auxiliary tasks with dynamic weights $\lambda_t$.
An hyperparameter $\rho$ as a scale factor is introduced to calculate the weight $\lambda_t$.  
Yin et al.~\cite{yin2018multi} proposed a multi-task model for face pose-invariant recognition with an automatic learning of the weights for each task. The main task of is set to 1 and the auxiliary tasks are sharing the dynamic tasks generated by the softmax layer. However, the update of the weights of tasks by the total loss of the networks runs counter to the objective of the multi-task learning. Thanks to the progress of the representation learning based on the deep neural networks, the methods based on the deep CNNs such as DeepFace~\cite{taigman2014deepface} , DeepIDs~\cite{sun2015deeply}, Facenet~\cite{schroff2015facenet}, VGGFace~\cite{simonyan2014very}, SphereFace~\cite{liu2017sphereface} have made a remarkable improvement comparing to the conventional methods based on the handcrafted features LBP, Gabor-LBP, HOG, SIFT~\cite{ahonen2006face, deniz2011face, bicego2006use, simonyan2013fisher}. The situation is same as facial expression recognition based on deep CNNs ~\cite{jung2015joint,zhao2016peak, mollahosseini2016going}. Even so, the studies on the face recognition with the facial expression images are limited. ~\cite{kakadiaris2007three, zhu2015high,chang2006multiple} propose the 3D based methods to deal with this issue. Kakadiaris et al.~\cite{kakadiaris2007three} present a fully automated framework for 3D face recognition using the Annotated Face Model to converted the raw image of face to a geometric model and a normal map. Then the face recognition is based on the processed image by using the Pyramid and Haar. Zhu et al.~\cite{zhu2015high} presents the method by using the 3D facial model to normalise the facial expression and then maps the normalised face to the 2D image to employ face recognition.  Chang et al.~\cite{chang2006multiple} describe a method using three different overlapping regions around the nose to employ the face recognition since this region is invariant in the presence of facial expression.

\section{Architecture}
The proposed multi-task learning with dynamic weights is based on the deep CNNs (see ~\figurename~\ref{fig:framework}). The hard parameter sharing structure is adopted as our framework, in which the sharing hidden layers are shared between all tasks~\cite{ruder2017overview}. The task-specific layers consisting of two branches are respectively dedicated to face verification and facial expression recognition. The two branches have almost identical structures facilitate the transfer learning of facial expression recognition from the pretrained face recognition task. Specifically, the BRANCH 1 can extract the embedded features of bottleneck layer for face verification and the BRANCH 2 uses the fully connected softmax layer to calculate the probabilities of the facial expressions. The deep CNNs in this work are based on the Inception-ResNet structure which have 13 million parameters of about 20 hidden layers in terms of the depth and 3 branches to the maximum in terms of the large. By the virtue of the Inception structure, the size of the parameters is much fewer than other popular deep CNNs such as VGGFace with ~138 million parameters.

\textbf{Dynamic-weight-unit} The dynamic weights of tasks are generated by the softmax layer connecting to the end of the sharing hidden layers, which can be so called the Dynamic-weight-unit. Each element in the Dynamic-weight-unit is corresponding to a weight of a task, thus the size of the Dynamic-weight-unit is equal to the number of weights of tasks, e.g. the size is 2 in this work. Since the weights are generated by the softmax layer,  $w1+w2=1$ which can well indicate the relative importance of the tasks. The parameters of this softmax layer are updated by the independent loss function $\mathcal{L}_3$ during the training of the networks,  which can automatically adjust the weights of tasks in light of the variation of the loss of tasks and drive the networks to always train the hard task firstly by assigning a larger weight.  


\section{Multi-task learning with dynamic weights}

The total loss of the proposed multi-task CNNs is the sum of the weighted losses of the multiple tasks.

(I) \noindent~\textbf{Multi-task loss $\mathcal{L}$}: The multi-task total loss $\mathcal{L}$ is defined as follows:
\begin{equation}
\label{eq1}
\mathcal{L}(\mathbf{X};\Theta;\Psi) = \sum_{i=1}^{T} w_i(\Psi)\mathcal{L}_i(\mathbf{X}_i;\Theta_i)
\end{equation}
where $T$ is the number of the tasks, here $T=2$. ${X_i}$ and ${\Theta_i}$ are the feature and the parameters corresponding to each task, $\Theta=\{\Theta_i\}_{i=1}^{T}$ are the overall parameters of the networks to be optimized by $\mathcal{L}$.  $\Psi$ are the parameters of the softmax layer in the Dynamic-weight-unit used to generate the dynamic weights  $w_i \in[0,1]$ s.t. $\sum w_i=1$. Thus $\{\mathbf{X}_i, \Theta_i\}\in \mathbb{R}^{d_i}$, where $d_i$ is the dimension of the features $X_i$, and $\{\mathcal{L}_i, w_i\}\in \mathbb{R}^1$.  Particularly, when $w_1$ = 1, $w_2$ = 0 the multi-task networks are degraded as the single-task networks  for face verification (i.e. Branch 1 and sharing hidden layers) while  $w_1$ = 0, $w_2$ = 1 is corresponding to the single-task networks for facial expression recognition (i.e. Branch 2 and sharing hidden layers). 

(II)\noindent~\textbf{Face verification task loss} $\mathcal{L}_1$: The loss for face verification task is measured by the center loss~\cite{wen2016discriminative} joint with the cross-entropy loss of softmax of Branch 1. The loss function of face verification task $\mathcal{L}_1$ is given by:      
\begin{equation}
\label{L1}
\mathcal{L}_1(\mathbf{X}_1;\Theta_1) = \mathcal{L}_{s1}(\mathbf{X}_1;\Theta_1) + \alpha\mathcal{L}_c(\mathbf{X}_1;\Theta_1)
\end{equation}

where $\mathcal{L}_{s1}$ is the cross-entropy loss of softmax of Branch 1, $\mathcal{L}_{c}$ is the center loss weighted by the hyperparameter $\alpha$. The $\mathcal{L}_{c}$ can be treated as a regularization item of softmax loss $\mathcal{L}_{s1}$ which is given by:
\begin{equation}
\label{eq:cross_entropy} 
\begin{split}
\mathcal{L}_{s1}(\mathbf{X}_1;\Theta_1)&= \sum^{K}_{k=1}-y_klogP(y_k=1|\mathbf{X}_1,\theta_k)\\  &=-\sum^{K}_{k=1}y_klog\frac{e^{f^{\theta_k}(\mathbf{X}_1)}}{\sum^{K}_{k'} e^{f^{\theta_{k'}}(\mathbf{X}_1)}}
\end{split}
\end{equation}

where $K$ is the number of the classes, i.e. the number of  identities in the training dataset, ${y_k}\in\{0,1\}$ is the one-shot label of the feature $\mathbf{X}_1$, $P(y_k|\mathbf{X}_1,\theta_k)$ is softmax function over the activation function $f^{\theta_k}(\mathbf{X}_1)$ where  $\{\theta_k\}_{k=1}^{K}=\Theta_1$, $\theta_k\in \mathbb{R}^{d_1}$. The bottleneck layer of BRANCH 1 is extracted as the feature $\mathbf{X}_1$ of the input image. The center loss $\mathcal{L}_c$ is given by:

\begin{equation}
\label{eq:centerloss} 
\mathcal{L}_c(\mathbf{X}_1;\Theta_1) = ||\mathbf{X}_1-C_{y_k}||
\end{equation}
Where the $C_{y_k}$ is the center of the class which $\mathbf{X}_1 $ belonging to, $C_{y_k} \in \mathbb{R}^{d_1}$.

(III)\noindent~\textbf{Facial expression recognition task loss $\mathcal{L}_2(\mathbf{X}_2;\Theta_2)$}: The loss function of facial expression recognition task  $\mathcal{L}_2$ is the cross-entropy loss of the softmax layer of BRANCH 2. The equation of $\mathcal{L}_2$ is as same as Equation~\ref{eq:cross_entropy} except the $K$ in $\mathcal{L}_{2}$ is the number of the categories of the facial expressions, $\mathbf{X}_2$ is the bottleneck layer of BRANCH 2, $\Theta2$ is corresponding parameters of this task.

(IV)\noindent~\textbf{Generation of the dynamic weights $w_i(\Psi)$}:
The dynamic weights $w_i$ are generated by the softmax layer of the dynamic-weight-unit which is given by: 
\begin{equation}
\label{eq1_dynamicweights}
w_i(\mathbf{Z};\Psi) = \frac{e^{f^{\psi_i}(\mathbf{Z})}}{\sum^{T}_{i'}e^{f^{\psi_{i'}}(\mathbf{Z})}}
\end{equation}
where the $\mathbf{Z} \in \mathbb{R}^{d_z}$ is the flat output of the last layer of the sharing hidden layers.$T$ is the number of the tasks, here $T$=2. $\psi_i$ is parameters in the softmax layer of the dynamic-weight-unit $\{\psi_i\}_{i=1}^{T}=\Psi$, $\psi_i\in \mathbb{R}^{d_z}$. $f^{\psi_{i}}(\mathbf{Z})$ is activation function which is given by:
\begin{equation}
\label{activefunction}
f^{\psi_i}(\mathbf{Z}) = \psi_i\mathbf{Z}^T + b_i
\end{equation}
Note that, we do not use the Relu function as the activation function since Relu discards the values minors zero. This shrinks the range of the variation of the dynamic weights $w_i$.
 
(V) \noindent~\textbf{Update of the dynamic weights $w_i$}:
Rather than using the total loss to update the dynamic weights, we propose a new loss function to update the dynamic weights which can drive the networks always train the hard task. The proposed new loss function for updating the dynamic weights is given by:
\begin{equation}
\label{eq:L3} 
\mathcal{L}_{3}(\mathbf{Z};\Psi)=\sum^{T}_{i=1}\frac{w_i(\psi_i)}{\mathcal{L}_i(\Theta_i)} \quad s.t. \quad \sum w_i=1
\end{equation}
Note that, $\mathcal{L}_i\{\Theta_i\}$ is independent with $w_i(\psi_i)$ since $\Theta_i\cap \psi_i=\varnothing$ , $i \in [1,..,T]$, thus $\mathcal{L}_i$ is constant for the dynamic weight update loss function $\mathcal{L}_3$. 

(VI) \noindent~\textbf{Qualitative analysis} shows that when the loss of the task $\mathcal{L}_i$ is small, i.e. the reciprocal of the $\mathcal{L}_i$ is large, thus loss $\mathcal{L}_3$ will try to reduce the loss by decreasing the value of $w_i$. That is to say, when the task is easy with a small loss the weight of the task will be assigned by a small value. On the contrary, the hard task with a large loss will be assigned by a large weight, which enable the networks always focus on training the hard task firstly. The update of the dynamic weights $w_i$ is essentially the update of the parameters $\psi_i$ which generate the dynamic weights. 

(VII) \noindent~\textbf{Quantitative analysis}:
Considering the \equationautorefname~\ref{eq1_dynamicweights} and \equationautorefname~\ref{activefunction}, the gradient of the $\psi_i$ can be given by
\begin{equation}
\label{eq:gradientofweights} 
\nabla{\psi_i}=\frac{\partial \mathcal{L}_3}{\partial \psi_i} =\frac{1}{\mathcal{L}_i}\frac{\partial w_i(\psi_i)}{\partial \psi_i}= \frac{1}{\mathcal{L}_i}\frac{a_i\sum^{T}_{j\neq i}a_{j}}{(\sum_i^T a_i)^2}\mathbf{Z}
\end{equation}
where $a_i = e^{{\psi_i}\mathbf{Z}^T+b_i}$, and the update of the parameters is  $\psi_i^{t+1}=\psi_i^{t}-\eta\nabla{\psi_i}^t$ where $\eta$ is the learning rate. Then the new value of the dynamic weight $w_i^{t+1}$ can be obtained by the Equation~\ref{eq1_dynamicweights} and ~\ref{activefunction} with the  $\psi_i^{t+1}$.

If we assume the $b_i^0=0, \psi_i^0=0$ (this is possible if we initialize the $\psi_i, b_i$ by zero), the $\psi_i^t$ can be given by
\begin{equation}
\label{eq:parametersupdate} 
\psi_i^t=-\sum\frac{1}{\mathcal{L}_i}\frac{a_i\sum^{T}_{j\neq i}a_{j}}{(\sum_i^T a_i)^2}\mathbf{Z}
\end{equation}
if we consider the case for two tasks $w_1$ and $w_2$ when $t=1$:
\begin{equation}
\label{eq:dynamicweightsanaylsis} 
\begin{split}
\frac{w_1^t}{w_2^t}&=e^{(\psi_1^t-\psi_2^t)\mathbf{Z}^T} \\
&=e^{(\frac{1}{\mathcal{L}_2}-\frac{1}{\mathcal{L}_1})\frac{a_1a_2}{(a_1+a_2)^2}\mathbf{Z}\mathbf{Z}^T}
\end{split}
\end{equation}
We can see that $a_i>0$ and $ZZ^T\ge 0$, so if $\mathcal{L}_2 < \mathcal{L}_1$ the $\frac{w_1}{w_2}>1$ namely $w_1>w_2$. It means if the loss of task1 larger than the loss of task 2, the weight of the task1 is larger than the one of task2. It indicates that the proposed loss function $\mathcal{L}_3$ can well update the weights of tasks to drive the networks always train the hard task firstly.


(VIII)\noindent~\textbf{Training protocol}:  The training of the entire deep CNNs includes two independent training: the training of the parameters of the networks {$\Theta$} by the multi-task loss $\mathcal{L}(\Theta)=\sum_{i=1}^{2}\mathcal{L}_i(\theta_i)$ and the training of the parameters of weight-generate-module {$\Psi$} by the loss $\mathcal{L}_3(\Psi)$. These  can be conducted simultaneously in a parallel way.
\begin{equation}
\Theta^{t-1}-\eta\frac{\partial\mathcal{L}(\Theta)}{\partial\Theta}\mapsto\Theta^{t}
\end{equation}
\begin{equation}
\Psi^{t-1}-\eta\frac{\partial\mathcal{L}_3(\Psi)}{\partial\Psi}\mapsto \Psi^{t}
\end{equation}
where $\eta\in(0,1)$ is the learning rate.
\section{Experiments and analysis}

~\subsection{Datasets}

\begin{figure}[t]
\begin{center}
  \includegraphics[width=0.72\linewidth]{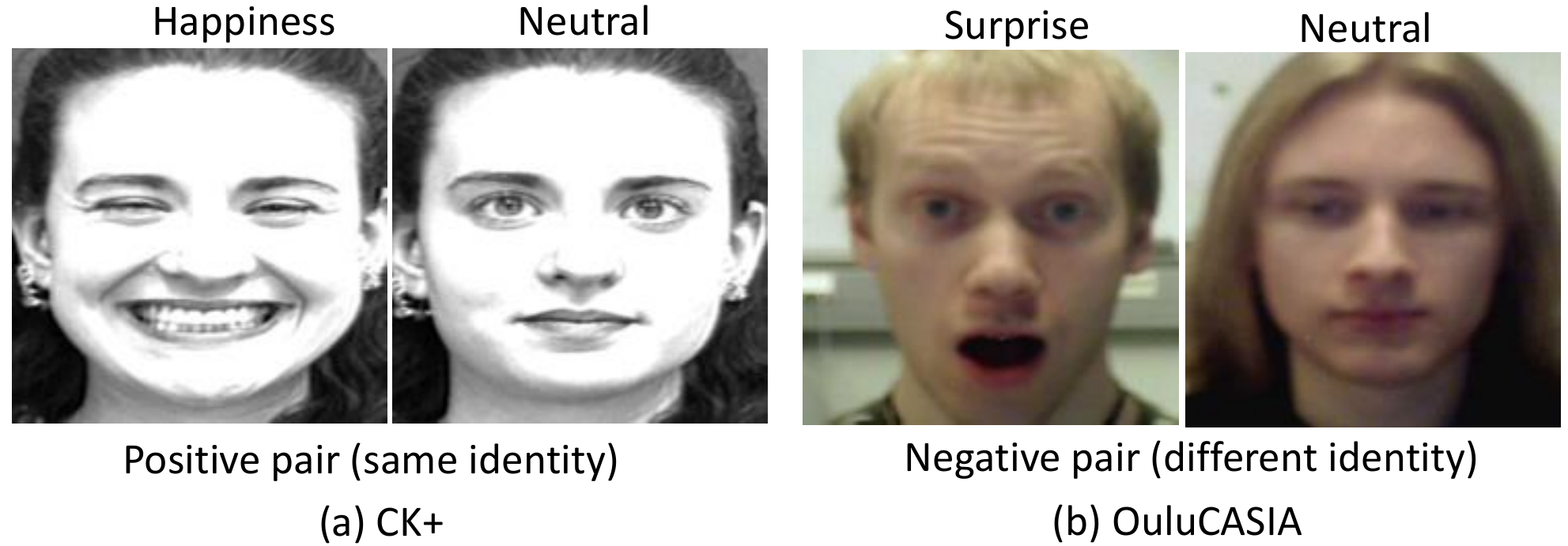}
\end{center}
  \caption{The image from CK+ and OuluCASIA.}
\label{fig:datasets}
\label{fig:onecol}
\end{figure}

Since the proposed multi-task networks performs the face verification task and the facial expression recognition task simultaneously, the datasets including both identity labels and facial expression labels are necessary to the training and the evaluation of the model. However, the large-scale datasets such as Celeb-A~\cite{liu2015faceattributes} and FER2013~\cite{goodfellow2013challenges} either do not include the facial expression or the identity labels.
Finally 5184 (positive or negative) pairs of face images with both identity labels and facial expression labels are extracted from OuluCASIA as well as 2099 pairs of images are extracted from CK+ to form two datasets respectively (see Fig.~\ref{fig:datasets} and Table~\ref{table1:expressiondatasets} in which ID (identities), Neutral (Ne), Anger (An), Disgust (Di), Fear (Fe), Happy (Ha), Sad (Sa), Surprise (Su), Contempt (Co).). 
 
 \setlength\tabcolsep{4 pt}
\begin{table}
\centering
\small
\caption{The datasets used in the multi-task learning for face verification and facial expression recognition in this work.}
\label{table1:expressiondatasets}
\begin{tabular}{@{}l|ccccccccc@{}}
\hline
\multicolumn{1}{c}{} & ID & Ne & An & Di & Fe & Ha & Sa & Su & Co  \\
\hline
CK+  & 123 & 327 & 135 &  177 & 75  & 147  & 84  & 249  & 54  \\
OuluCASIA & 560& 560   & 240 & 240  &  240& 240  & 240  & 240  & -  \\
\hline

\end{tabular}
\end{table}

\subsection{Experimental configuration}  The faces have been detected by the MTCNN~\cite{zhang2016joint} from the raw images. The RMSprop with the mini-batches of 90 samples are applied for optimizing the parameters. The learning rate is started from 0.1, and decay by 10 at the different iterations. The networks are initialized by Xavier~\cite{glorot2010understanding} and biases values are set to zero at beginning. The momentum coefficient is set to 0.99.  The dropout with the probability of 0.5 and the weight decay of 5e-5 are applied.  The weight of the center loss $\alpha$ is set to 1e-4.

~\subsection{Pretrained model}
Before the training of the proposed multi-task CNNs, a single-task network constituted of the sharing hidden layers and the BRANCH 1 is pretrained for face verification-task with large-scale dataset by loss function $\mathcal{L}_1$. Then the training of the dynamic multi-task CNNs can handling on the pretrained model. Moreover, in order to compare the multi-task learning with the single-task learning, the BRANCH 2 is also trained independently by transferring the learning of the pretrained BRANCH 1 for facial expression recognition with loss function $\mathcal{L}_2$. Finally we obtain two models pretrained by the single-task learning for face verification (sharing layers + BRANCH 1) and facial expression recognition (sharing layers + BRANCH 2) respectively.

\subsection{Dynamic multi-task learning / training}
\begin{figure*}
\centering
\begin{minipage}{.24\textwidth}
  \centering
  \includegraphics[width=\linewidth]{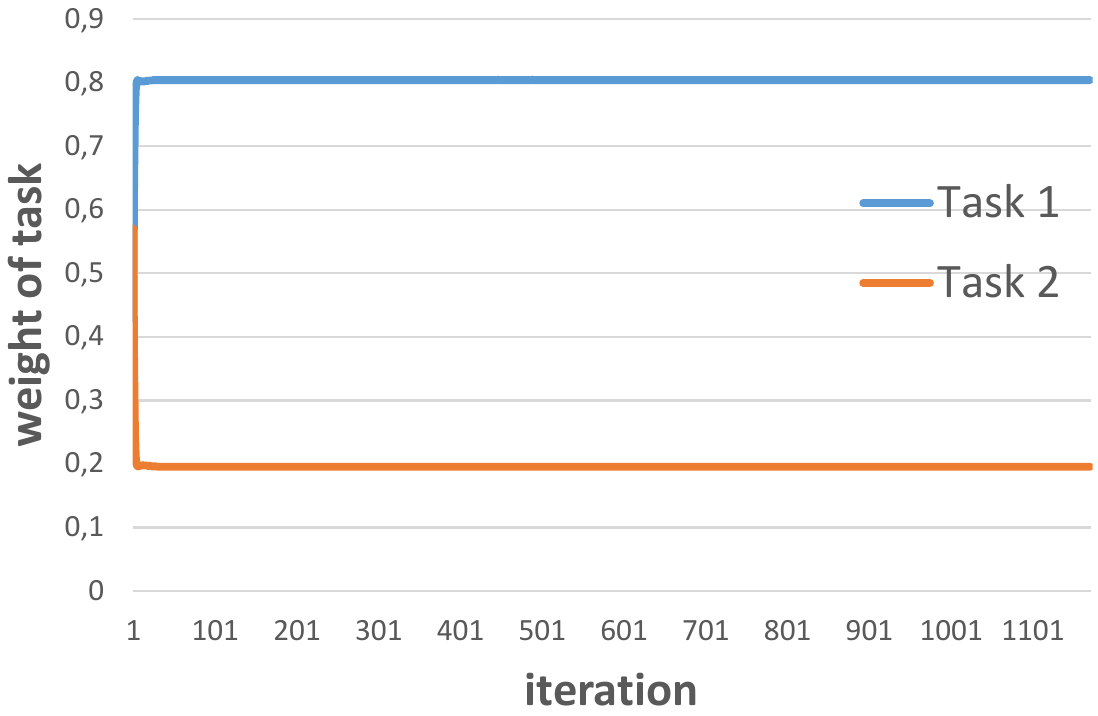}
  \subcaption{Naive dynamic - CK+}
  \label{}
\end{minipage}
\begin{minipage}{.24\textwidth}
  \centering
  \includegraphics[width=\linewidth]{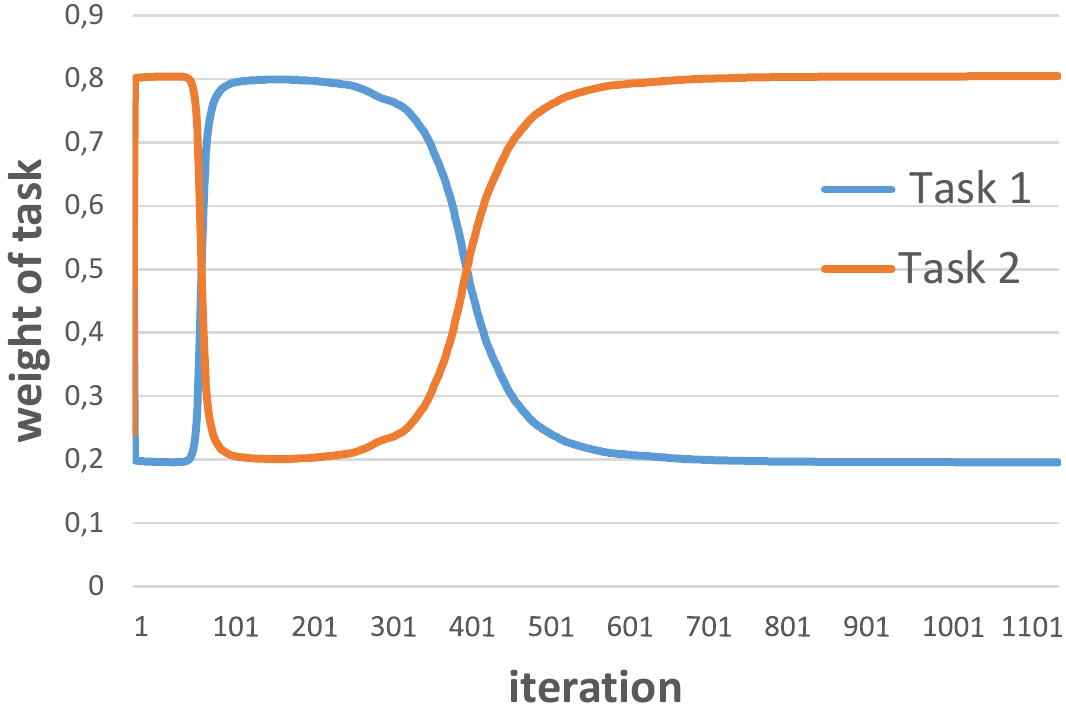}
  \subcaption{Proposed dynamic - CK+}
  \label{fig:dynweights_real_CK+}
\end{minipage}
\begin{minipage}{.24\textwidth}
  \centering
  \includegraphics[width=\linewidth]{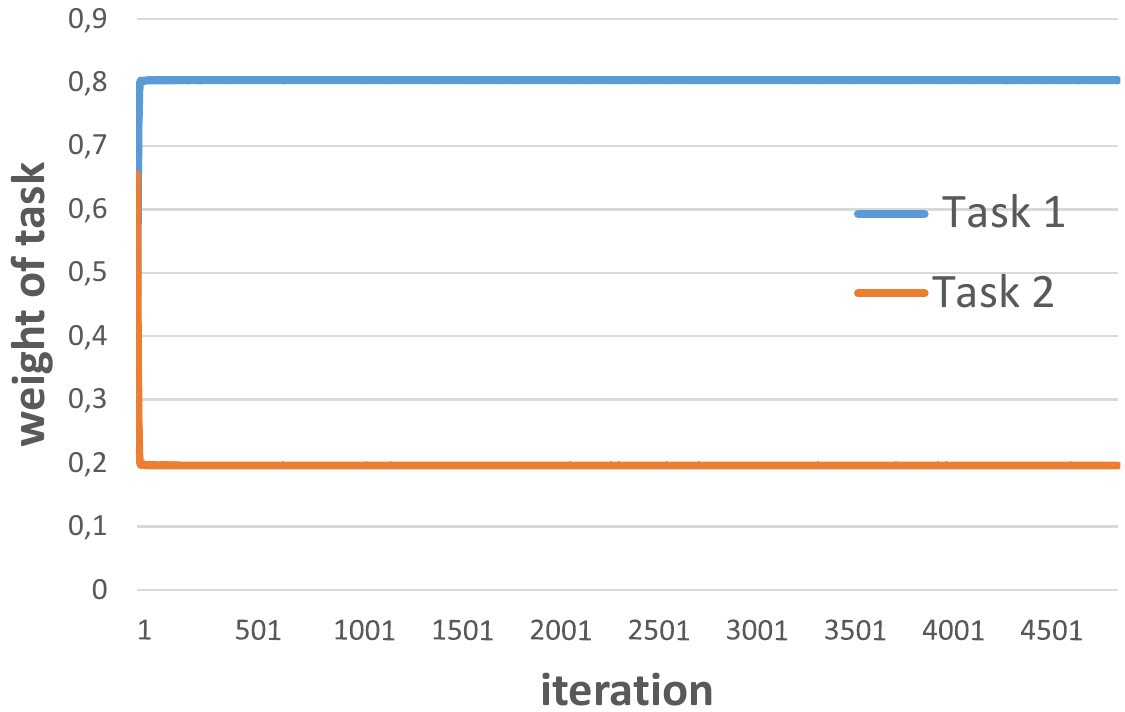}
  \subcaption{Naive dynamic - Oulu}
  \label{fig:naivedynweights_Oulu}
\end{minipage}
\begin{minipage}{.24\textwidth}
  \centering
  \includegraphics[width=\linewidth]{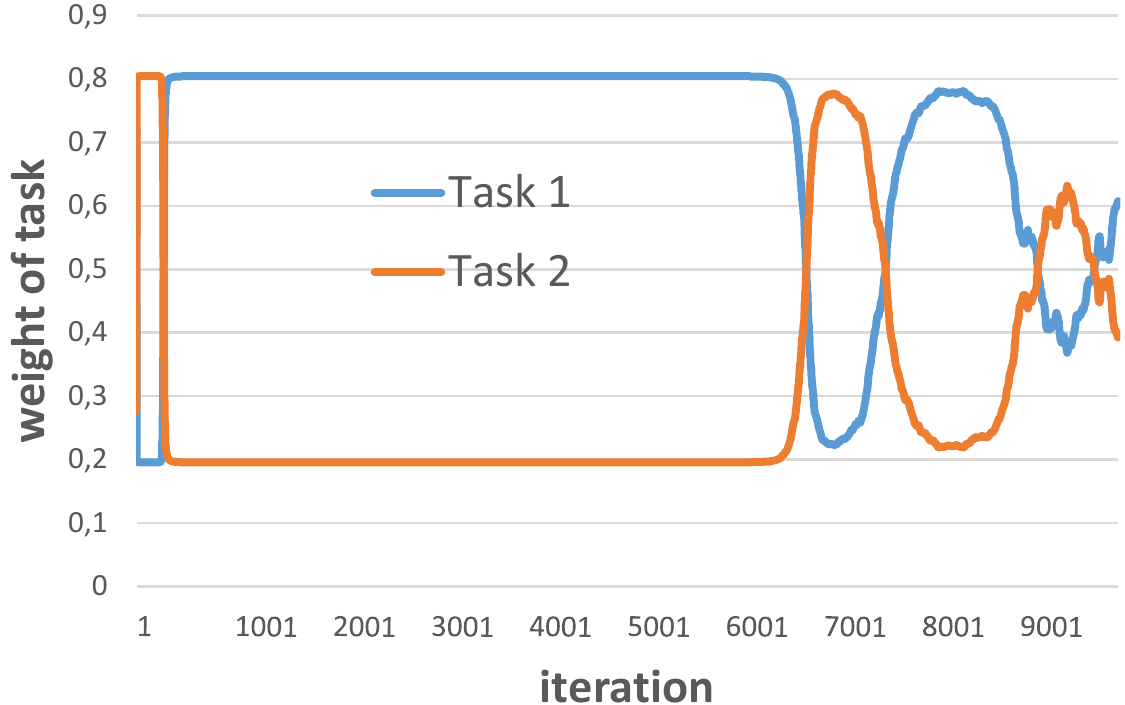}
  \subcaption{Proposed dynamic - Oulu}
  \label{fig:dynweights_oulu}
\end{minipage}
\caption{The weights of tasks generated by our proposed method and the naive dynamic method during the training. Task 1 is facial expression recognition and Task 2 is face verification.}
\label{fig:full-dynamic-weights}
\end{figure*}

\begin{figure*}
\centering
\begin{minipage}{.24\textwidth}
  \centering
  \includegraphics[width=\linewidth]{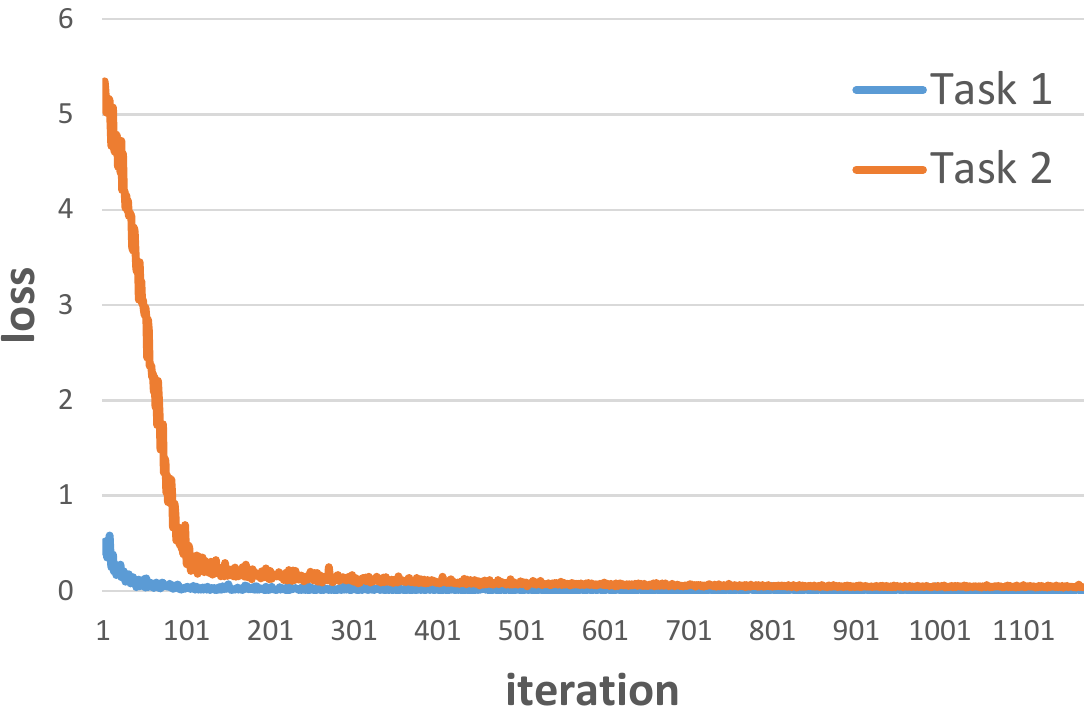}
  \subcaption{Naive dynamic - CK+}
  \label{fig:naviedynloss_CK+}
\end{minipage}%
\begin{minipage}{.24\textwidth}
  \centering
  \includegraphics[width=\linewidth]{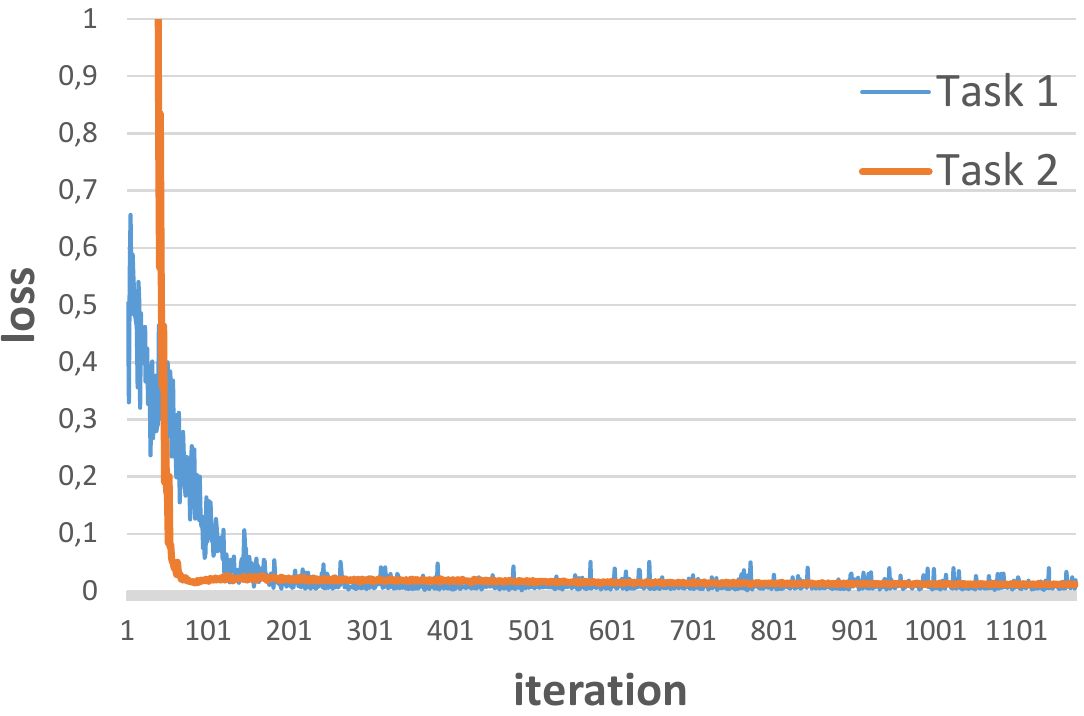}
  \subcaption{Proposed dynamic - CK+}
  \label{fig:dynloss_real_CK+}
\end{minipage}
\begin{minipage}{.24\textwidth}
  \centering
  \includegraphics[width=\linewidth]{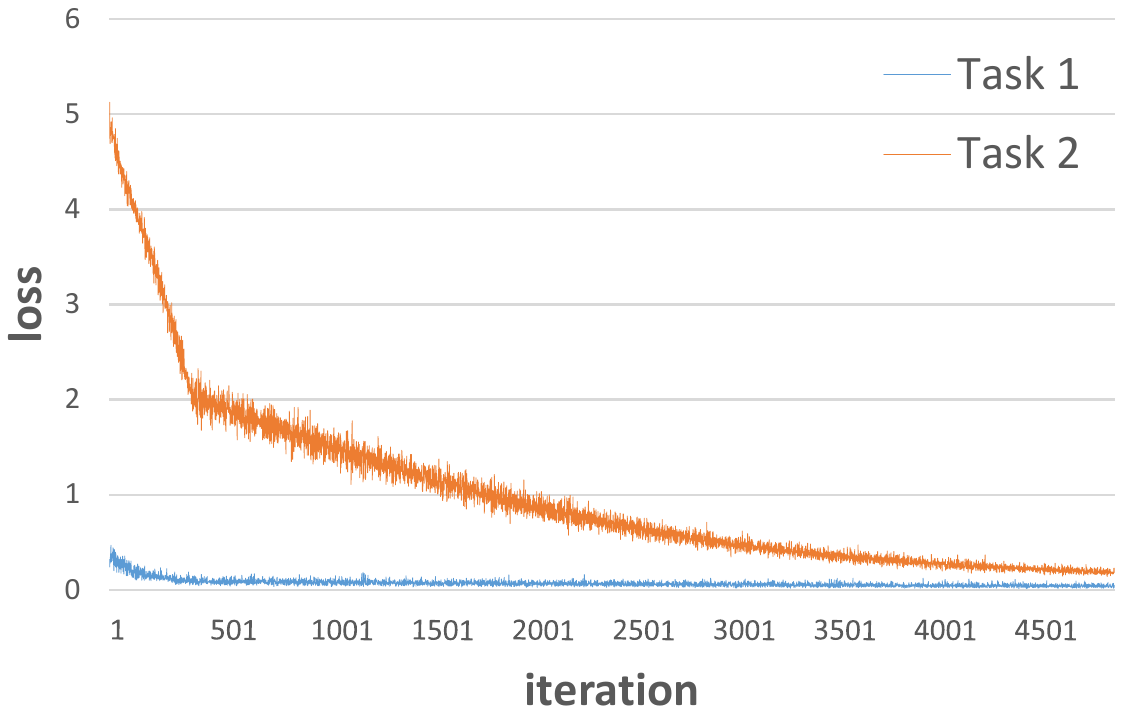}
  \subcaption{Naive dynamic - Oulu}
  \label{fig:naivedynloss_Oulu}
\end{minipage}
\begin{minipage}{.24\textwidth}
  \centering
  \includegraphics[width=\linewidth]{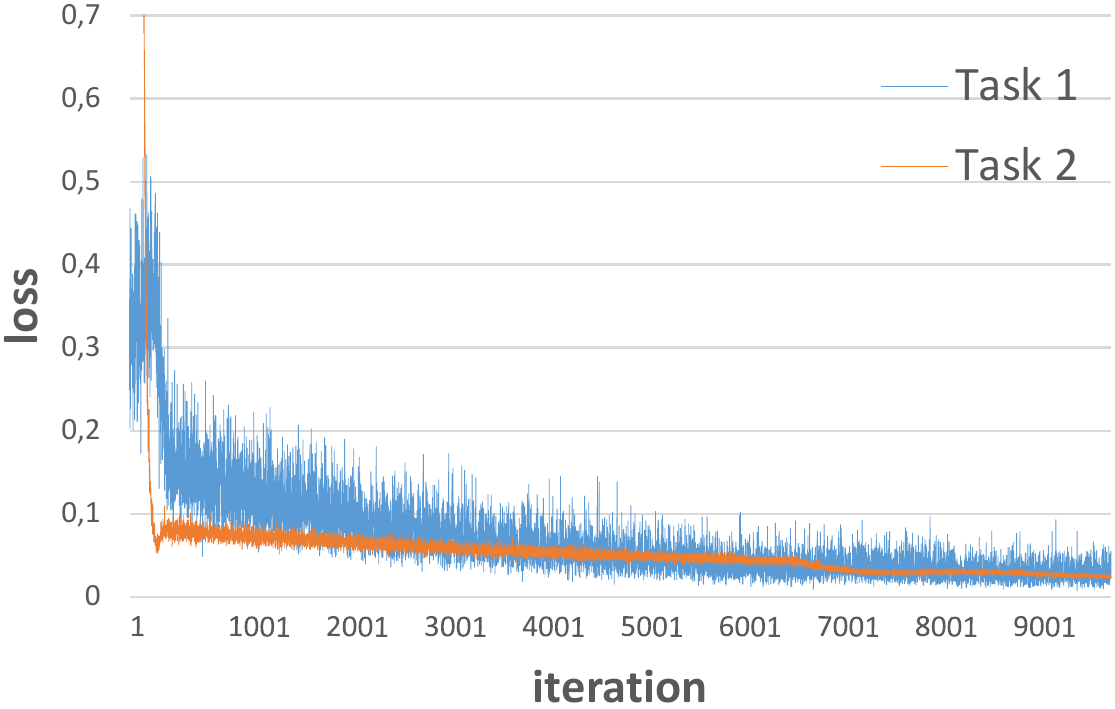}
  \subcaption{Proposed dynamic - Oulu}
  \label{fig:dynloss_oulu}
\end{minipage}
\caption{The loss of tasks corresponding to our proposed method and the naive dynamic method during the training. Task 1 is facial expression recognition and Task 2 is face verification.}
\label{fig:full-dynamic-loss}
\end{figure*}

\begin{figure*}
\centering
\begin{minipage}{.24\textwidth}
  \centering
  \includegraphics[width=\linewidth]{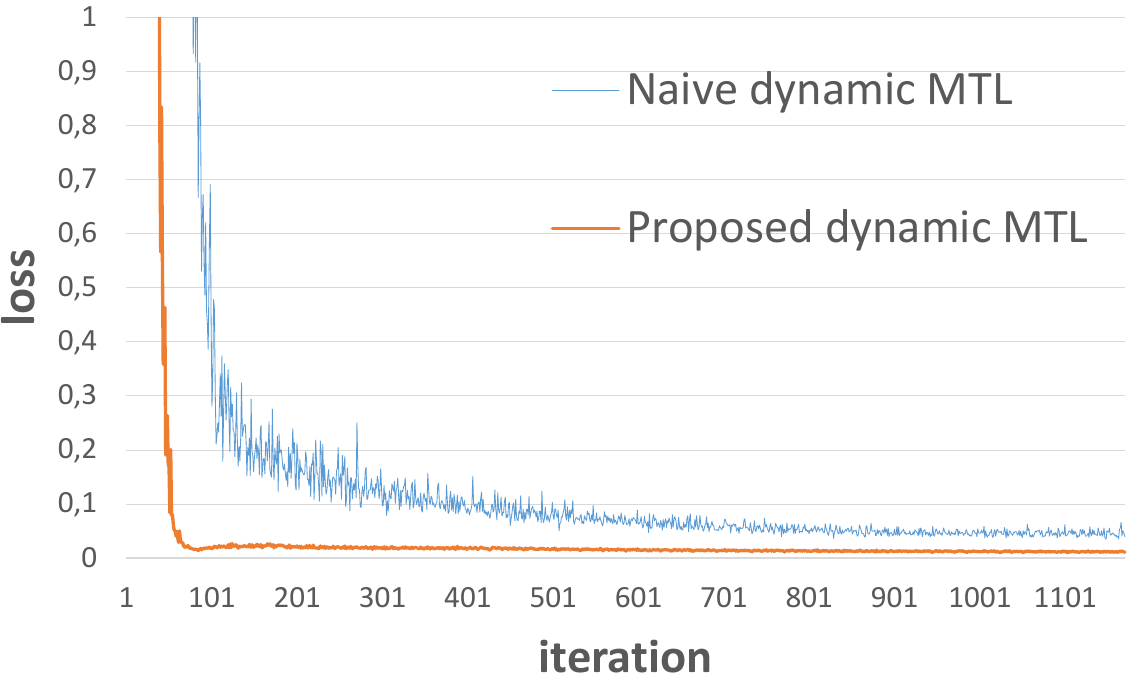}
  \subcaption{Face verification - CK+}
  \label{fig:Fig5_lossvs_verif_CK+}
\end{minipage}
\begin{minipage}{.24\textwidth}
  \centering
  \includegraphics[width=\linewidth]{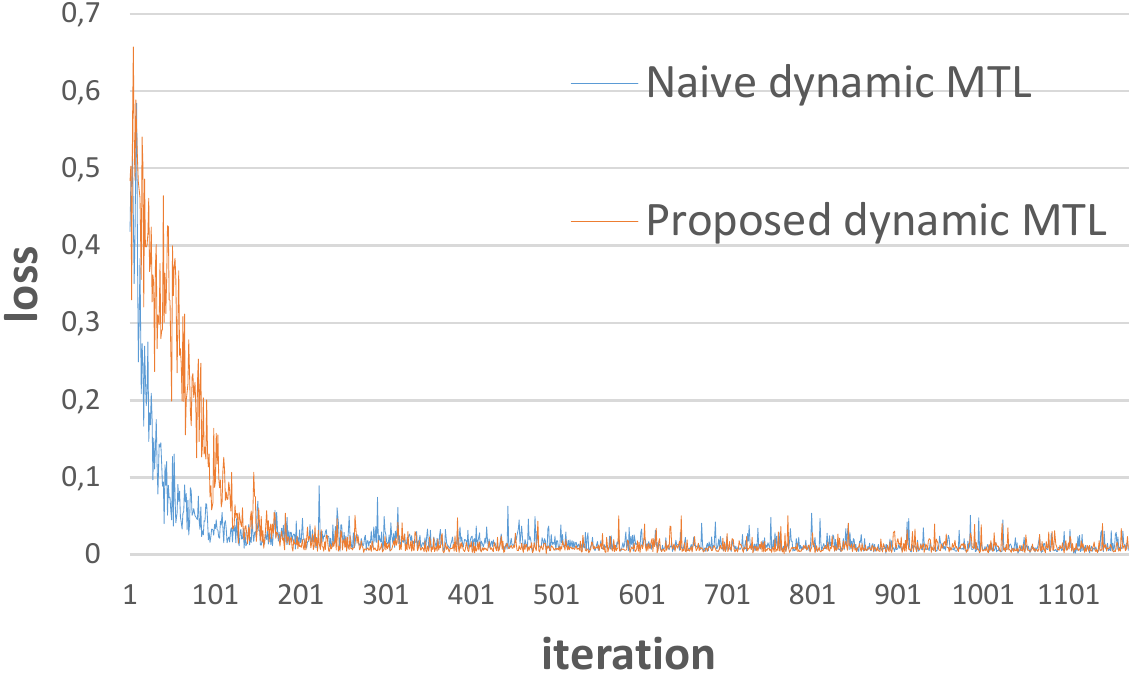}
  \subcaption{Facial expression - CK+}
  \label{fig:lossvs_expr_CK+}
\end{minipage}
\begin{minipage}{.25\textwidth}
  \centering
  \includegraphics[width=\linewidth]{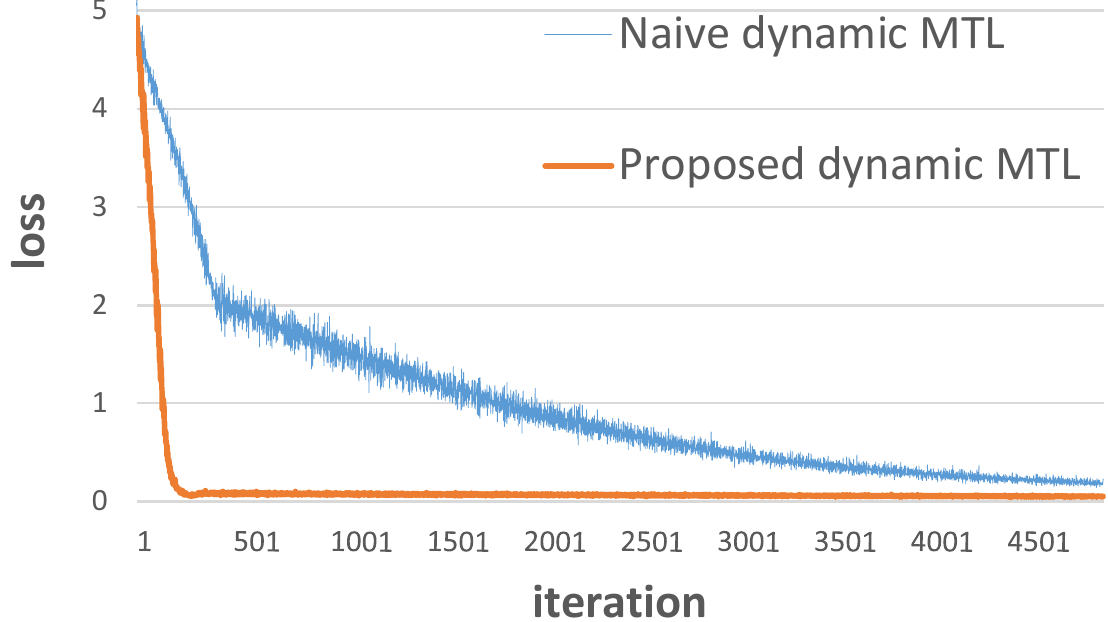}
  \subcaption{Face verification - Oulu}
  \label{fig:lossvs_verif_oulu_cropped}
\end{minipage}
\begin{minipage}{.23\textwidth}
  \centering
  \includegraphics[width=\linewidth]{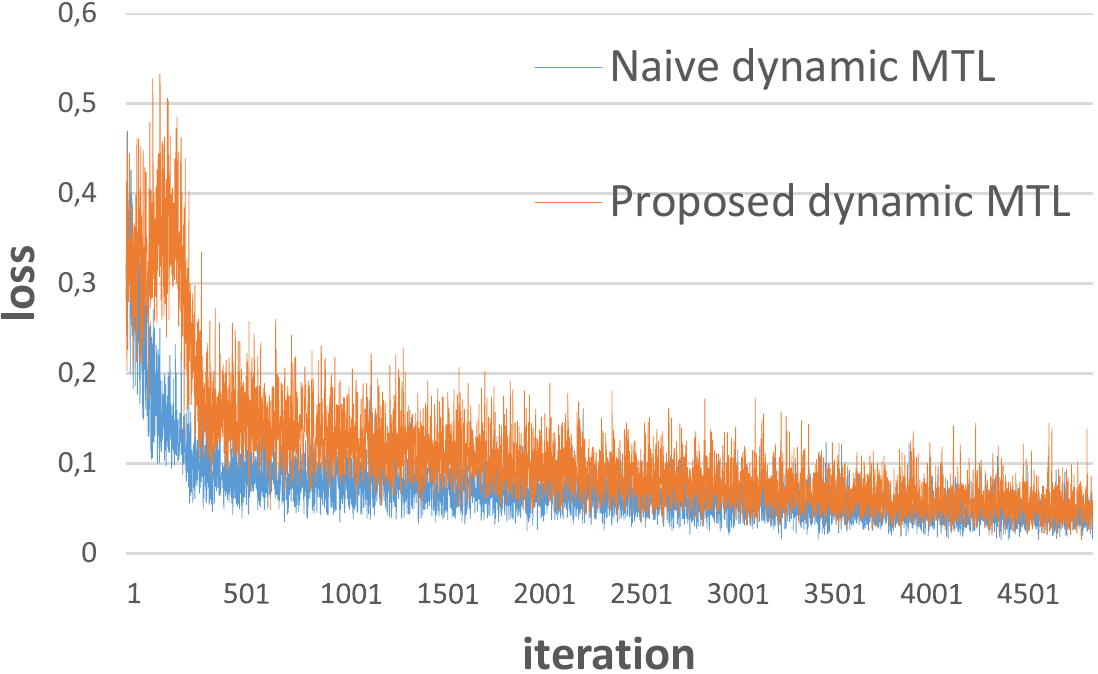}
  \subcaption{Facial expression - Oulu}
  \label{fig:lossvs_expr_oulu}
\end{minipage}
\caption{The training efficiency for decreasing the loss of task by the proposed dynamic multi-learning method and the naive dynamic multi-learning method on different tasks, i.e. face verification and facial expression recognition.}
\label{fig:Fig5_lossvs}
\end{figure*}

In order to distinguish our proposed method, we call the method in~\cite{yin2018multi} as naive dynamic method. Comparing to the naive dynamic method, the proposed dynamic method can adjust the weights of tasks according to their importance/training difficulty as shown in \figurename~\ref{fig:full-dynamic-weights}. The training difficulty of the task is presented by its training loss. \figurename~\ref{fig:full-dynamic-loss} shows the variation of the loss of tasks corresponding to the two different methods. From \figurename~\ref{fig:full-dynamic-weights} and \figurename~\ref{fig:full-dynamic-loss}, we can see that the naive dynamic method always train the easy task namely facial expression recognition (denoted as Task 1) with smaller loss by assigning a large weight as shown in (a) on dataset CK+ or (c) on dataset OuluCASIA . However, the hard task namely face verification (denoted as Task 2) with large loss is always assigned by small weight less than 0.2. Contrarily, the weight of task generated by the proposed method can effectively adapt to the varied importance of the task in the multi-task learning. For instance, as shown in the (b) on dataset CK+, the hard task which is face verification (Task 2) with a large loss is assigned a large weight at the beginning of the training. The large weight of task drive the networks to fully train the hard task so that the loss of the hard task decreases rapidly and soon it is lower than the loss of the task of facial expression recognition (Task 1). Once the previous easy task become the hard task with a larger loss, the proposed method automatically assigns a larger weight to the current hard task as shown in (b) that the weight of the facial expression recognition (Task 1) augment promptly from the bottom to the top when the loss of the task becomes the larger one. Thus the networks are capable to switch to fully train the current hard task with the proposed dynamic method. \figurename~\ref{fig:Fig5_lossvs} 
suggests that the proposed dynamic method can decrease the loss of the hard task, i.e. the face verification task, more quickly and achieve lower value of loss. For the easy task, namely the facial expression recognition task, these two methods decrease the loss similarly since the easy task can be sufficiently trained by both of the methods. Thus the proposed dynamic method is superior to the naive dynamic method in terms of the training efficiency. 

\begin{figure*}
\centering
\begin{minipage}{.24\textwidth}
  \centering
  \includegraphics[width=\linewidth]{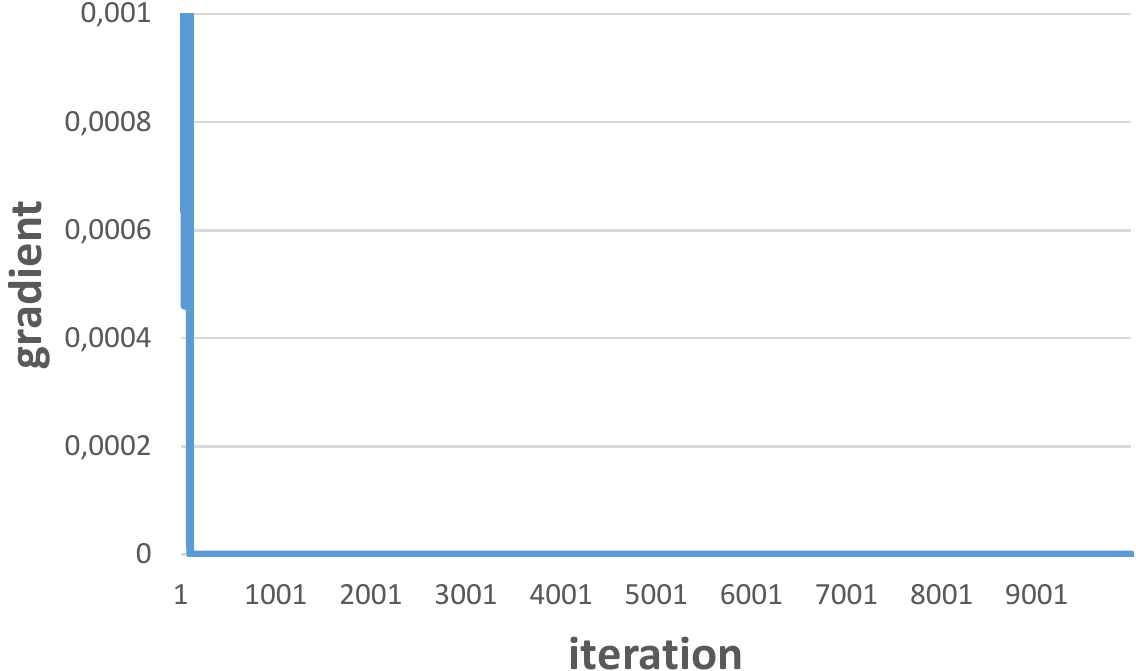}
  \subcaption{Original gradient}
  \label{fig:Fig6_lossvs_verif_CK+}
\end{minipage}%
\begin{minipage}{.24\textwidth}
  \centering
  \includegraphics[width=\linewidth]{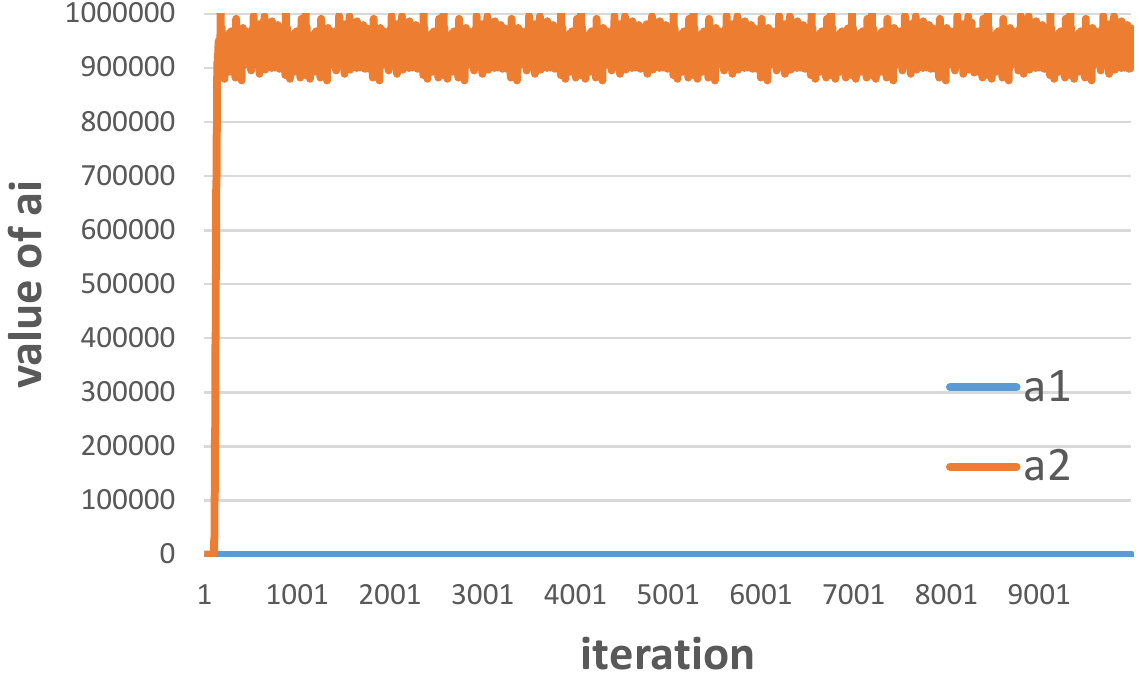}
  \subcaption{Original $a_i$}
  \label{fig:grad_vanishing_ai}
\end{minipage}
\begin{minipage}{.21\textwidth}
  \centering
  \includegraphics[width=\linewidth]{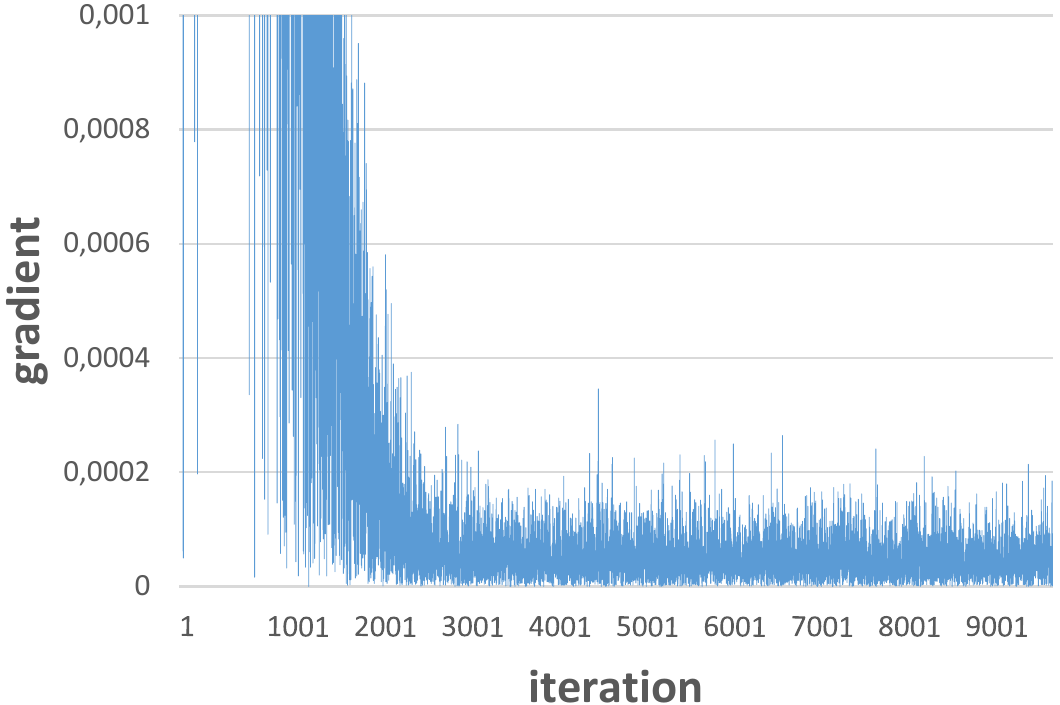}
  \subcaption{Normalized gradient}
  \label{fig:grad_embed}
\end{minipage}
\begin{minipage}{.24\textwidth}
  \centering
  \includegraphics[width=\linewidth]{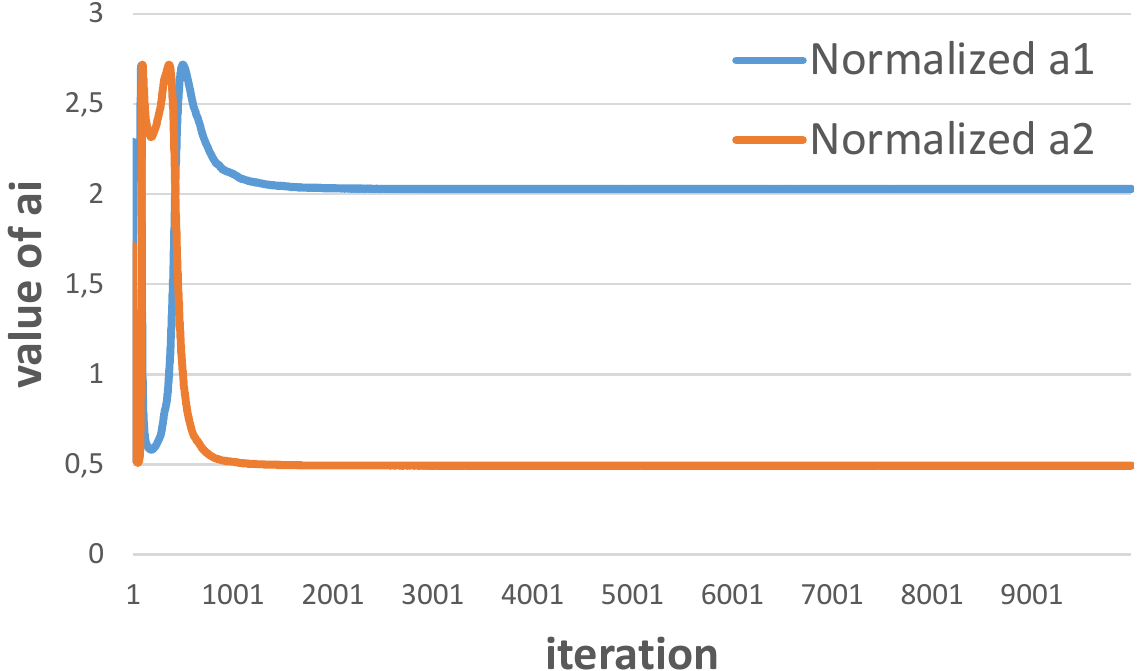}
  \subcaption{Normalized $a_i$}
  \label{fig:grad_embed_ai}
\end{minipage}
\caption{The gradient value before and after the normalization of $a_i$. The normalization of the $a_i$ can mitigate the gradient vanishing problem caused by the large value of $a_i$.}
\label{fig:gradvanish}
\end{figure*} 

\subsection{Gradient vanishing problem}
In this section we analyse the problem of the gradient vanishing for updating the dynamic weights. From \equationautorefname~\ref{eq:gradientofweights}, we can see that if the $a_i\sum^{T}_{j\neq i}a_{j}<<{(\sum_i^T a_i)^2}$, the $\nabla{\psi_i} \rightarrow{} 0$ which means that the gradient vanishes. In this work, $T=2$, if $0<<a_1^2+a_2^2+a_1a_2$, it will cause the problem of the gradient vanishing. Since the $a_i = e^{{\psi_i}\mathbf{Z}^T+b_i}>0$, the condition of gradient vanishing is easy to satisfy provided the $a_i$ is relative large. In order to mitigate the problem of grand vanishing, we normalize the $a_i$ as the embedding feature for calculating the weights. As shown in (a) and (b) of \figurename~\ref{fig:gradvanish}, the gradient vanishes when the $a_i$ is large than $8*10^6$. By applying the normalization of $a_i$ as shown in (d). the gradient return to the normal values as shown in (c).

\subsection{Evaluation and ablation analysis}
~\textbf{(I) Dynamic multi-task learning for face verification with facial expression}
To evaluate the effectiveness of the proposed dynamic multi-task learning method for face verification with facial expression, we firstly analyse the results from the single-task method with pretrained models trained on the general datasets and then the fine-tuning model based on the datasets used in this work. Furthermore, we compare the multi-task learning methods with manually setting weights (i.e. static multi-task learning), naive dynamic weights and our proposed dynamic weights to the single-task learning method. 

Table~\ref{tab:multitaskeval} firstly shows that the performance of the state-of-art methods such as DeepID, VGGFace, FaceNet, etc. with pretrained models for face verification with facial expression. Comparing to the performance on the general dataset such as LFW or YTF, we can see that the performances on the face images with facial expression in CK+ or OuluCASIA have degraded obviously, e.g. the face verification accuracy of DeepID has decreased from 99.47\% on LFW to 91.70\% on CK+, VGGFace has decreased from 98.95\% on LFW to 92.20\% on CK+ as well as FaceNet which is trained on the large scale dataset. 
This is quite probably resulted by the lack of the facial expression images in the general datasets for the training of the models. By fine-tuning our pretrained model with the facial expression datasets, the performance has improved. 
Thanks to the capacity of learning the features between the tasks, the multi-task learning with fixed weights further improve the performance comparing to the fine-tuning single task model.
However, the performance of face verification with the naive dynamic multi-task learning is inferior to the static multi-task learning and even the fine-tuning single-task model, which is due to the face verification task 
has not been sufficiently trained with a small weight. 
Rather than the static multi-task learning method with the fix weights or the naive dynamic multi-task learning, the proposed dynamic method can real-time update the weights of tasks according to the importance of tasks and achieve the best performance. 
 
 ~\textbf{(II) Dynamic multi-task learning for facial expression recognition}
Table~\ref{tab:accexpressionck+} and Table~\ref{tab:accexpressionOulu} compare the proposed dynamic multi-task learning for facial expression recognition with other methods on CK+ and OuluCASIA respectively. As well as the face verification task, the proposed dynamic multi-task learning achieves the best performance on both datasets. Since the facial expression recognition is the easy task,  the naive dynamic multi-task learning has sufficiently trained this task and achieved the comparable results as the proposed method. The multi-task learning also show the significant improvement to the single-task methods.

  \begin{table}
\caption{\label{tab:multitaskeval}The evaluation of face verification on facial expressions datasets with different methods (accuracy\%).}
\begin{center}
\small
\begin{tabular}{|l|c|c|c|c|c|}
\hline
Method & Images & LFW & YTF & CK+ & Oulu.\\
\hline\hline
DeepFace~\cite{taigman2014deepface} &4M&${97.35}$&${91.4}$ &-&-  \\\hline
DeepID-2,3~\cite{sun2015deeply} &-&${99.47}$&${93.2}$&91.70&96.50  \\\hline
FaceNet~\cite{schroff2015facenet} &200M&\textbf{99.63}&\textbf{95.1}&98.00&97.50  \\\hline
VGGFace~\cite{simonyan2014very} &2.6M&${98.95}$&${91.6}$&92.20&93.50  \\\hline
Centerloss~\cite{wen2016discriminative} &0.7M&${99.28}$&${94.9}$&94.00&95.10  \\\hline
SphereFace~\cite{liu2017sphereface} &0.7M&${99.42}$&${95.0}$&93.80&95.50  \\\hline
{Single-task (pretrained)}  & 1.1M & 99.41& 95.0&98.00&92.60 \\
{Single-task (fine-tuning)}  & 1.1M & 99.10 & 94.2 &98.50&97.71 \\
{Static MTL}  & 1.1M &99.23 &94.1 &98.50&98.00 \\
{Naive dynamic MTL}  & 1.1M &99.23 &94.1 &98.15&95.14 \\
{Proposed dynamic MTL}  & 1.1M & 99.21 & 94.3&\textbf{99.00}&\textbf{99.14} \\\hline
\end{tabular}
\end{center}
\end{table}

\begin{table}
\begin{center}
\small
\caption{\label{tab:accexpressionck+}The evaluation of proposed multi-task networks for facial expression recognition task on dataset CK+.}
    \begin{tabular}{|l|c|}
      \hline
      Method & Accuracy(\%) \\
      \hline\hline
      LBPSVM~\cite{feng2007facial}&${95.1}$   \\\hline
      Inception~\cite{mollahosseini2016going} &${93.2}$  \\\hline
      DTAGN~\cite{jung2015joint} &${97.3}$ \\\hline
      PPDN~\cite{zhao2016peak}&${97.3}$   \\\hline
      AUDN~\cite{liu2013aware} &${92.1}$   \\\hline
      Single-task &${98.21}$   \\
      Static MTL &${99.11}$   \\
      Naive dynamic MTL &${99.10}$   \\
      Proposed dynamic MTL &\textbf{99.50}\\\hline   
      \end{tabular}
\end{center}
\end{table}

\begin{table}
\begin{center}
\small
\caption{\label{tab:accexpressionOulu}The evaluation of proposed multi-task networks for facial expression recognition task on dataset OuluCASIA.}
	\begin{tabular}{|l|c|}
	\hline
  Method & Accuracy(\%) \\
  \hline\hline
  HOG3D~\cite{klaser2008spatio}&${70.63}$   \\\hline
  AdaLBP~\cite{zhao2011facial}&${73.54}$ \\\hline
  DTAGN~\cite{jung2015joint} &${81.46}$ \\\hline
  PPDN~\cite{zhao2016peak}&${84.59}$   \\\hline
  Single-task &85.42   \\
  Static MTL &\textbf{89.60}   \\
  Naive dynamic MTL &88.89   \\
  Proposed dynamic MTL &\textbf{89.60}\\\hline   
  \end{tabular}
\end{center}
\end{table}

\section{Conclusion}
In this work, we propose a dynamic multi-task learning method which allows to dynamically update the  weight of task according to the importance of the task during the training process. Comparing to the other multi-task learning methods, our method does not introduce the hyperparameters and it enables the networks to focus on the training of the hard tasks which results a higher efficiency and better performance for training the multi-task learning networks. Either the theoretical analysis or the experimental results demonstrate the effectiveness of our method. 
This method can be also easily applied in the other deep multi-task learning frameworks such as Faster R-CNN for object detection.

\section*{Acknowledgment}
This work was supported by the MOBIDEM project, part of the ``Systematic Paris-Region'' and ``Images \& Network'' Clusters, funded by the French government. 

{\small
\bibliographystyle{ieee}
\bibliography{egbib_final}

\begin{thebibliography}{10}\itemsep=-1pt

\bibitem{ahonen2006face}
T.~Ahonen, A.~Hadid, and M.~Pietikainen.
\newblock Face description with local binary patterns: Application to face
  recognition.
\newblock {\em IEEE transactions on pattern analysis and machine intelligence},
  28(12):2037--2041, 2006.

\bibitem{bicego2006use}
M.~Bicego, A.~Lagorio, E.~Grosso, and M.~Tistarelli.
\newblock On the use of sift features for face authentication.
\newblock In {\em Computer Vision and Pattern Recognition Workshop, 2006.
  CVPRW'06. Conference on}, pages 35--35. IEEE, 2006.

\bibitem{chang2006multiple}
K.~I. Chang, K.~W. Bowyer, and P.~J. Flynn.
\newblock Multiple nose region matching for 3d face recognition under varying
  facial expression.
\newblock {\em IEEE Transactions on Pattern Analysis and Machine Intelligence},
  28(10):1695--1700, 2006.

\bibitem{chen2017multi}
W.~Chen, X.~Chen, J.~Zhang, and K.~Huang.
\newblock A multi-task deep network for person re-identification.
\newblock In {\em AAAI}, pages 3988--3994, 2017.

\bibitem{chen2017gradnorm}
Z.~Chen, V.~Badrinarayanan, C.-Y. Lee, and A.~Rabinovich.
\newblock Gradnorm: Gradient normalization for adaptive loss balancing in deep
  multitask networks.
\newblock {\em arXiv preprint arXiv:1711.02257}, 2017.

\bibitem{collobert2008unified}
R.~Collobert and J.~Weston.
\newblock A unified architecture for natural language processing: Deep neural
  networks with multitask learning.
\newblock In {\em Proceedings of the 25th international conference on Machine
  learning}, pages 160--167. ACM, 2008.

\bibitem{deng2013new}
L.~Deng, G.~Hinton, and B.~Kingsbury.
\newblock New types of deep neural network learning for speech recognition and
  related applications: An overview.
\newblock In {\em Acoustics, Speech and Signal Processing (ICASSP), 2013 IEEE
  International Conference on}, pages 8599--8603. IEEE, 2013.

\bibitem{deniz2011face}
O.~D{\'e}niz, G.~Bueno, J.~Salido, and F.~De~la Torre.
\newblock Face recognition using histograms of oriented gradients.
\newblock {\em Pattern Recognition Letters}, 32(12):1598--1603, 2011.

\bibitem{feng2007facial}
X.~Feng, M.~Pietik{\"a}inen, and A.~Hadid.
\newblock Facial expression recognition based on local binary patterns.
\newblock {\em Pattern Recognition and Image Analysis}, 17(4):592--598, 2007.

\bibitem{girshick2015fast}
R.~Girshick.
\newblock Fast r-cnn.
\newblock In {\em Proceedings of the IEEE international conference on computer
  vision}, pages 1440--1448, 2015.

\bibitem{glorot2010understanding}
X.~Glorot and Y.~Bengio.
\newblock Understanding the difficulty of training deep feedforward neural
  networks.
\newblock In {\em 13th International Conference on Artificial Intelligence and
  Statistics}, pages 249--256, 2010.

\bibitem{goodfellow2013challenges}
I.~J. Goodfellow, D.~Erhan, and et~al.
\newblock Challenges in representation learning: A report on three machine
  learning contests.
\newblock In {\em International Conference on Neural Information Processing},
  pages 117--124, 2013.

\bibitem{huang2007labeled}
G.~B. Huang, M.~Ramesh, T.~Berg, and E.~Learned-Miller.
\newblock Labeled faces in the wild: A database for studying face recognition
  in unconstrained environments.
\newblock Technical report, Technical Report 07-49, University of
  Massachusetts, Amherst, 2007.

\bibitem{jung2015joint}
H.~Jung, S.~Lee, and et~al.
\newblock Joint fine-tuning in deep neural networks for facial expression
  recognition.
\newblock In {\em Proceedings of the IEEE International Conference on Computer
  Vision}, pages 2983--2991, 2015.

\bibitem{kakadiaris2007three}
I.~A. Kakadiaris, G.~Passalis, G.~Toderici, M.~N. Murtuza, Y.~Lu,
  N.~Karampatziakis, and T.~Theoharis.
\newblock Three-dimensional face recognition in the presence of facial
  expressions: An annotated deformable model approach.
\newblock {\em IEEE Transactions on Pattern Analysis and Machine Intelligence},
  29(4):640--649, 2007.

\bibitem{kendall2018multi}
A.~Kendall, Y.~Gal, and R.~Cipolla.
\newblock Multi-task learning using uncertainty to weigh losses for scene
  geometry and semantics.
\newblock In {\em Proceedings of the IEEE Conference on Computer Vision and
  Pattern Recognition}, pages 7482--7491, 2018.

\bibitem{klaser2008spatio}
A.~Klaser, M.~Marsza{\l}ek, and C.~Schmid.
\newblock A spatio-temporal descriptor based on 3d-gradients.
\newblock In {\em BMVC 2008-19th British Machine Vision Conference}, pages
  275--1. British Machine Vision Association, 2008.

\bibitem{liu2013aware}
M.~Liu, S.~Li, and et~al.
\newblock Au-aware deep networks for facial expression recognition.
\newblock In {\em Automatic Face and Gesture Recognition, 2013 IEEE
  International Conference and Workshops on}, pages 1--6. IEEE, 2013.

\bibitem{liu2017sphereface}
W.~Liu, Y.~Wen, Z.~Yu, M.~Li, B.~Raj, and L.~Song.
\newblock Sphereface: Deep hypersphere embedding for face recognition.
\newblock In {\em The CVPR}, volume~1, page~1, 2017.

\bibitem{liu2015faceattributes}
Z.~Liu, P.~Luo, X.~Wang, and X.~Tang.
\newblock Deep learning face attributes in the wild.
\newblock In {\em Proceedings of International Conference on Computer Vision
  (ICCV)}, 2015.

\bibitem{lucey2010extended}
P.~Lucey, J.~F. Cohn, and et~al.
\newblock The extended cohn-kanade dataset (ck+): A complete dataset for action
  unit and emotion-specified expression.
\newblock In {\em CVPR Workshops, 2010 IEEE Computer Society Conference on},
  pages 94--101. IEEE, 2010.

\bibitem{mollahosseini2016going}
A.~Mollahosseini, D.~Chan, and et~al.
\newblock Going deeper in facial expression recognition using deep neural
  networks.
\newblock In {\em Applications of Computer Vision, 2016 IEEE Winter Conference
  on}, pages 1--10. IEEE, 2016.

\bibitem{murugesan2016adaptive}
K.~Murugesan, H.~Liu, J.~Carbonell, and Y.~Yang.
\newblock Adaptive smoothed online multi-task learning.
\newblock In {\em Advances in Neural Information Processing Systems}, pages
  4296--4304, 2016.

\bibitem{parkhi2015deep}
O.~M. Parkhi, A.~Vedaldi, A.~Zisserman, et~al.
\newblock Deep face recognition.
\newblock In {\em BMVC}, page~6, 2015.

\bibitem{ranjan2017hyperface}
R.~Ranjan, V.~M. Patel, and R.~Chellappa.
\newblock Hyperface: A deep multi-task learning framework for face detection,
  landmark localization, pose estimation, and gender recognition.
\newblock {\em IEEE Transactions on Pattern Analysis and Machine Intelligence},
  2017.

\bibitem{ruder2017overview}
S.~Ruder.
\newblock An overview of multi-task learning in deep neural networks.
\newblock {\em arXiv preprint arXiv:1706.05098}, 2017.

\bibitem{schroff2015facenet}
F.~Schroff, D.~Kalenichenko, and J.~Philbin.
\newblock Facenet: A unified embedding for face recognition and clustering.
\newblock In {\em CVPR}, pages 815--823, 2015.

\bibitem{simonyan2013fisher}
K.~Simonyan and O.~M. e.~a. Parkhi.
\newblock Fisher vector faces in the wild.
\newblock In {\em BMVC}, page~4, 2013.

\bibitem{simonyan2014very}
K.~Simonyan and A.~Zisserman.
\newblock Very deep convolutional networks for large-scale image recognition.
\newblock {\em arXiv preprint arXiv:1409.1556}, 2014.

\bibitem{sun2015deeply}
Y.~Sun, X.~Wang, and X.~Tang.
\newblock Deeply learned face representations are sparse, selective, and
  robust.
\newblock In {\em CVPR}, pages 2892--2900, 2015.

\bibitem{taigman2014deepface}
Y.~Taigman, M.~Yang, and et~al.
\newblock Deepface: Closing the gap to human-level performance in face
  verification.
\newblock In {\em CVPR}, pages 1701--1708, 2014.

\bibitem{tian2015pedestrian}
Y.~Tian, P.~Luo, X.~Wang, and X.~Tang.
\newblock Pedestrian detection aided by deep learning semantic tasks.
\newblock In {\em Proceedings of the CVPR}, pages 5079--5087, 2015.

\bibitem{wen2016discriminative}
Y.~Wen, K.~Zhang, Z.~Li, and Y.~Qiao.
\newblock A discriminative feature learning approach for deep face recognition.
\newblock In {\em European Conference on Computer Vision}, pages 499--515.
  Springer, 2016.

\bibitem{wolf2011face}
L.~Wolf, T.~Hassner, and I.~Maoz.
\newblock Face recognition in unconstrained videos with matched background
  similarity.
\newblock In {\em CVPR, 2011 IEEE Conference on}, pages 529--534. IEEE, 2011.

\bibitem{yim2015rotating}
J.~Yim, H.~Jung, B.~Yoo, C.~Choi, D.~Park, and J.~Kim.
\newblock Rotating your face using multi-task deep neural network.
\newblock In {\em Proceedings of the CVPR}, pages 676--684, 2015.

\bibitem{yin2018multi}
X.~Yin and X.~Liu.
\newblock Multi-task convolutional neural network for pose-invariant face
  recognition.
\newblock {\em IEEE Transactions on Image Processing}, 27(2):964--975, 2018.

\bibitem{zhang2016joint}
K.~Zhang, Z.~Zhang, Z.~Li, and Y.~Qiao.
\newblock Joint face detection and alignment using multitask cascaded
  convolutional networks.
\newblock {\em Signal Processing Letters}, 23(10):1499--1503, 2016.

\bibitem{zhang2016learning}
Z.~Zhang, P.~Luo, C.~C. Loy, and X.~Tang.
\newblock Learning deep representation for face alignment with auxiliary
  attributes.
\newblock {\em IEEE transactions on pattern analysis and machine intelligence},
  38(5):918--930, 2016.

\bibitem{zhao2011facial}
G.~Zhao, X.~Huang, and et~al.
\newblock Facial expression recognition from near-infrared videos.
\newblock {\em Image and Vision Computing}, 29(9):607--619, 2011.

\bibitem{zhao2016peak}
X.~Zhao, X.~Liang, and et~al.
\newblock Peak-piloted deep network for facial expression recognition.
\newblock In {\em European Conference on Computer Vision}, pages 425--442.
  Springer, 2016.

\bibitem{zhu2015high}
X.~Zhu, Z.~Lei, J.~Yan, D.~Yi, and S.~Z. Li.
\newblock High-fidelity pose and expression normalization for face recognition
  in the wild.
\newblock In {\em Proceedings of the IEEE Conference on Computer Vision and
  Pattern Recognition}, pages 787--796, 2015.

\end{thebibliography}
}
\end{document}